\documentclass[sigconf,natbib=true,anonymous=false]{acmart}
\usepackage{array}
\begin{document}

\title{Span-Aggregatable, Contextualized Word Embeddings for Effective Phrase Mining}

\author{Eyal Orbach}
\email{eyal.orbach@genesys.com}
\orcid{0009-0004-8689-2130}
\affiliation{%
  \institution{Genesys}
  \streetaddress{Menachem Begin, 146}
  \city{Tel Aviv}
  \country{Israel}
  \postcode{6492103}
}

\author{Lev Haikin}
\email{lev.haikin@genesys.com}
\affiliation{%
  \institution{Genesys}
  \streetaddress{Menachem Begin, 146}
  \city{Tel Aviv}
  \country{Israel}
  \postcode{6492103}
}

\author{Nelly David}
\email{nelly.david@genesys.com}
\affiliation{%
  \institution{Genesys}
  \streetaddress{Menachem Begin, 146}
  \city{Tel Aviv}
  \country{Israel}
  \postcode{6492103}
}

\author{Avi Faizakof}
\email{avi.faizakof@genesys.com}
\affiliation{%
  \institution{Genesys}
  \streetaddress{Menachem Begin, 146}
  \city{Tel Aviv}
  \country{Israel}
  \postcode{6492103}
}



\begin{abstract}
Dense vector representations for sentences made significant progress in recent years as can be seen on sentence similarity tasks. Real-world phrase retrieval applications, on the other hand, still encounter challenges for effective use of dense representations. We show that when target phrases reside inside noisy context, representing the full sentence with a single dense vector, is not sufficient for effective phrase retrieval. We therefore look into the notion of representing multiple, sub-sentence, consecutive word spans, each with its own dense vector. We show that this technique is much more effective for phrase mining, yet requires considerable compute to obtain useful span representations. Accordingly, we make an argument for contextualized word/token embeddings that can be aggregated for arbitrary word spans while maintaining the span's semantic meaning. We introduce a modification to the common contrastive loss used for sentence embeddings that encourages word embeddings to have this property. To demonstrate the effect of this method we present a dataset based on the STS-B dataset with additional generated text, that requires finding the best matching paraphrase residing in a larger context and report the degree of similarity to the origin phrase. We demonstrate on this dataset, how our proposed method can achieve better results without significant increase to compute.
\end{abstract}

\maketitle

\section{Introduction}
Many real world applications benefit from enabling a user to supply  origin phrases and having the system find paraphrases for these, inside a large corpus of text. Such application include sifting through contact center conversations, legal documents, issue tracking systems and more. Currently, such systems commonly rely on sparse representations such as BM25 \cite{bm25} and query expansion variations \cite{splade, elastic-elser}. Meanwhile, recent years introduced works like SBERT \citep{sentencebert} that have improved dense representations for sentences as commonly demonstrated in sentence similarity tasks like STS\citep{cer-stsb-2017-semeval}.

\begin{figure}
\centering
\begin{tabular}{| m{1.71cm}| m{1.71cm} | m{1.8em}| m{2.7cm}|}
\hline
\tiny{\textbf{Origin Phrase}} & \tiny{\textbf{Target Phrase}} & \tiny{\textbf{Simarity Degree}} & \tiny{\textbf{Target Phrase in Context}} \\
\hline
\tiny{A group of men play soccer on the beach.} &  \tiny{A group of boys are playing soccer on the beach.} & \tiny{3.6} & \tiny{Catching a glimpse of the ocean from the balcony, I noticed \textbf{a group of boys are playing soccer on the beach}, their laughter and shouts echoing up to my ears.} \\
\hline
\tiny{Blue and red plane in mid-air flight.} & \tiny{A blue and red airplane while in flight}. & \tiny{4.8} & \tiny{Fascinated by the view from his window seat, Frank couldn't help but stare at the vast expanse of clouds below him, until his attention was suddenly drawn to \textbf{a blue and red airplane while in flight}, soaring gracefully through the sky.} \\
\hline
\end{tabular}
\caption{Examples from STS-B-Context. Target phrase are in bold for readability, actual dataset does not include any such markers.}
\label{fig:stscontext-example}
\end{figure}

However, when real-world applications attempt to retrieve paraphrases, these target phrases commonly reside in a surrounding noisy context, which is not well accounted for in sentence similarity tasks. As these sentence representations encode an entire word sequence (e.g. sentence) into a single vector they tend to be greatly affected by noise, hindering the ability to use dense vectors when searching for sub-sentence phrases. This setting of phrase retrieval is still under-explored lacking both efficient ways to use dense vector representations as well as standard benchmarks to asses their quality. Existing public datasets provide either gold-label similarity degree \cite{cer-stsb-2017-semeval, bird-naacl2019} or a surrounding context for phrases \cite{pic-phrase-in-context-dataset, chen-2017-wikipedia-open-domain} but not both. In this work we share findings that contributed to a real-world application that requires finding paraphrases that reside inside longer contexts, and report their the degree of similarity (semantic distance \citep{mohammad-semantic-distance}) to enable the user with threshold tuning. As the real-world data cannot be shared publicly, to demonstrate our findings in a reproducible manner we introduce a modified version of STS-B \cite{cer-stsb-2017-semeval} with auto-generated context \footnote{Full dataset: https://anonymous.4open.science/r/stsb-context-test-E21A/sts-test-sentences-clean.tsv} as seen in the examples in Fig \ref{fig:stscontext-example}. 

We first look into two simple alternatives to utilize dense vector representations:
\begin{enumerate}
\renewcommand{\theenumi}{\Alph{enumi}}
\renewcommand{\labelenumi}{\theenumi)}
    \item\label{setup-a} A single dense vector representation for the entire context (i.e., the context being a longer sentence comprising of the target phrase surrounded by the generated text).
    \item\label{setup-b} A dense vector representation for any existing consecutive words span (i.e., n-gram, limited by a hyperparameter of maximum ngram size), in the context.
\end{enumerate}
As the setup \ref{setup-b} proves to have superior results, we take a further look into its compute requirements. Producing an embedding vector for each span, by inserting it as isolated input into a model, produces best results in our comparison, yet requires significant compute. 
Therefore we make an argument for contextualized embeddings that can be aggregated for arbitrary spans while maintaining each span's semantic meaning. We present a method to train a model to produce such token embeddings, which we call SLICE (Span-aggregated Late Interaction Contextualized Embeddings). The method is a modification to the loss function commonly used in contrastive learning for text similarity \citep{sentencebert}. We show that training with this method, even on a noisy labeled dataset, produces a model that outperforms equivalent methods of similar compute utilization.

our contributions therefore are:
\renewcommand{\theenumi}{\roman{enumi}}
\renewcommand{\labelenumi}{\theenumi)}
\begin{enumerate}
    \item We show the advantages of contextualized embeddings that can be aggregated for arbitrary spans, when mining for phrases. 
    \item We introduce a modified loss function that helps produce such token embeddings, train a model on a noisy labeled dataset, and show it outperforms equivalent methods of similar compute utilization.
    \item We introduce a dataset to evaluate mining for phrases with semantic distance scores between origin phrase and best matched phrase in context.
\end{enumerate}

\section{Related Work}
The setting of Phrase/Paraphrase Mining, especially under a definition of phrases that extends beyond bi-grams and noun-phrases, has garnered limited attention in research literature, yet some related tasks have contributed to the method we propose in this paper. We focus on dense vectors for word sequences, with separation of the representation of the query from the representation of the searchable context. We view cross-encoder techniques such as \citep{bert-cross-encoding}, as outside the scope of this work, due to their excessive compute requirements of reprocessing the corpus for each query (origin phrase).

\subsection{Phrase Similarity and Word Embeddings Composition}
\citet{yu-ettinger-2020-assessing} have demonstrated that averaging the word embeddings produced by common contextualized word embeddings models like BERT \citep{bert} does not yield good representations for phrases or sentences. Following this, the work of SBERT \citep{sentencebert} showed how an appropriate training objective, actually can produces effective sentence embedding when applying mean-pooling on the output embeddings of BERT. Employing a contrastive learning approach, with triplets of sentences from Textual Entailment datasets, the training aims to push the sentence embedding of matching sentences (i.e. entailment) to be closer to each other and for random pairs to be less similar. Following the success of this approach, more models were trained with annotated datasets such as  MS-MARCO \cite{msmarco}. \citet{phrasebert} produce a corpus of paraphrases by mining phrases from Wikipedia and auto-generating paraphrases to train PhraseBERT. While they experiment with a loss function that also considers the context as input, it is not designed to separate the representation of the context from the phrase, and their evaluation tests the phrases in isolation. 
\subsection{Document Search}
Many existing works on semantic search focus on document retrieval where a user expects ranking of the most relevant documents relating to a provided query. \citet{colbert} introduced ColBERT, an effective dense retrieval method with late interaction, that finds semantic matching between words in the query to words in the searched documents. \emph{Document Search} tasks learn to rank as opposed to a normalized semantic distance, and have no explicit constraints for requiring consecutive word spans matches in the retrieved documents. In spite of these limitation, we claim that \emph{Document Search} datasets, specifically MS-MARCO Passage Ranking \citep{msmarco}, can be viewed as a noisy training dataset, hypothesizing that many of the labeled matching documents, do contain a consecutive words span that is highly similar to the search query. We use MS-MARCO as a noisy training set. We also build on the paradigm of \emph{late interaction} presented in ColBERT \cite{colbert} but modify it to be constrained by consecutive word spans, and be agnostic to span length, adhering better to paraphrase matching.

\subsection{Question Answering}
The tasks of extractive question answering can also be viewed as a form of phrase retrieval, as it requires a model to return a phrase in the text that answers the given question. There are many differences between the settings of \emph{Question Answering} and \emph{Phrase Mining}, yet the work of \citet{lee-etal-2021-learning-dense} makes an observation we find useful. In tackling a large corpus where the answer is assumed to be some span in the text (requiring a higher resolution then complete sentences), they propose to represent all existing phrases in the corpus, where phrase means any n-gram of words from single word to MAX (specifically 20) words. In this paper we show the usefulness of this 'brute force' choice for representing the searchable corpus. \citet{lee-etal-2021-learning-dense} choose to minimize the space required to represent all existing spans by representing a span only with its start-token and end-token. Their suggested DensePhrases encoder, makes use of this choice
by learning to encode each word as a start-span and end-span vector. We test DensePhrases for the phrase mining task with the appropriate modification of encoding the query by the same encoder used for the phrases. Our experiments suggest that this choice to minimize the space requirements hinders the representation resulting in limited ability for finer grained similarity assessment. 

\section{Method}
We start with the following observation: when mining for sub-sentence phrases surround by longer contexts, existing sentence-encoders yield much better results when representing all existing spans in the searchable context. We limit the spans we look into by hyperparameters of MIN\_SIZE and MAX\_SIZE. Heuristics for selecting these hyperparameters can generally be derived from a sample of origin phrases or modified according to new origin phrases at the cost of reprocessing the searchable corpus. The values we selected for these experiments are held constant across the different models and setups.
As we show in Table \ref{tab:stsvontext-corelation} when comparing the first setup  \emph{``Full Context"} to the second \emph{``N-grams - Forward per N-gram"}, all evaluated models make significant gains when encoding each of the n-grams in the searchable context and selecting the n-gram with the embedding that is most similar (highest cosine similarity) to the source phrase. While this approach is highly effective it comes at significant cost in terms of compute, as it requires a forward pass of the model for each such n-gram in the searchable text. We therefore present a more compute optimized approach which can be seen in Table \ref{tab:stsvontext-corelation} on the third setup \emph{``N-grams - single forward pass''}. In this setup we produce token embeddings by a single forward pass of the entire sentence, and aggregate them by mean-pooling, for each span between MIN\_SIZE and MAX\_SIZE. 
In the next section we describe our SLICE method that learns to produce token embeddings that are optimized to support this property of being dynamically aggregatable.
\subsection{SLICE}
We aim for a model that processes a given context (e.g. sentence, passage) in a single forward pass, and produces word/sub-word embeddings with the following properties:
\begin{enumerate}
    \item Contextualized embeddings, in the sense that they represent the word's meaning in the specific context it is in.
    \item Span-aggregatable, meaning that mean-pooling of the word embeddings of an arbitrary span, results in a vector that represents the meaning of the span.
\end{enumerate}
To produce such a model we follow the optimization objective of \emph{late interaction}, where in contrast to the contrastive learning objective of SBERT \citep{sentencebert}, only a subset of the encoder's output logits are used to propagate loss. Specifically we maximize the similarity between the query to its best match span in the positive example. and minimize similarity with the best match span in the negative example.

Formally, given a dataset of triplets as so:
\begin{enumerate}
    \item\label{formalation-q} A phrase to be searched, (denoted $q$)
    \item A short passage, (denoted $p\_true$) containing a phrase that is semantically similar to $q$.
    \item A negative example passage (denoted $p\_false$) that does not include a phrase that is semantically similar to $q$.
\end{enumerate}

We denote $W$ as our encoder. Given a sequence of words (referring interchangeably also to sub-words), the encoder, produces a vector for each of the input sequence's words:
\[W(<w_1,w_2,…,w_n>)=<\vec{v_1},\vec{v_2},…,\vec{v_n}>\]

Let $S$ be a sequence of words $<w_1,w_2,…,w_n>$, we denote $a$ to be a predefined minimum span length and $b$ the predefined maximal span length, for example, $a = 1$ and $b = 20$.
We denote $spans(S)$ to be the operation that produces all consecutive n-grams at a length between $a$ and $b$.
\begin{multline*}
spans(S) = \{<w_i,w_{i+1},…,w_{i+k-1}> | \\ 
a \leq k \leq b,  1 \leq i \leq (n-k+1)\}
\end{multline*}    
We denote $\lvert\lvert\vec{v} \cdot \vec{u}\rvert\rvert $ to be the cosine distance metric normalized to $[0,1]$ range for simplification. 

Let $sim\_true$ be the maximal similarity value found between a span in $p\_true$ to $q$.
\begin{multline*}
sim\_true=max_{x\in{spans(p\_true)}}\\\lvert\lvert avg\_pooling(W(q))\cdot avg\_pooling(W(x))\rvert\rvert
\end{multline*} 
And respectively $sim\_false$ the maximal similarity value found between a span in $p\_false$ to $q$
\begin{multline*}
sim\_false=max_{x\in{spans(p\_false)}}\\\lvert\lvert avg\_pooling(W(q))\cdot avg\_pooling(W(x))\rvert\rvert
\end{multline*} 
We therefore wish to maximize $sim\_true$ with respect to $sim\_false$.  
This can be expressed in several ways and we choose a cross entropy loss, with a constant ($\lambda=30$), to ease convergence, which can be simplified as so:

\[L= -\lambda sim\_true + log(e^{\lambda sim\_true}+ e^{\lambda sim\_false})\]

$L$ is fully differentiable, and therefore can be used to optimize the parameters of our encoder model $W$. 
We train an encoder based on BERT \cite{bert}, initialized with its pretrained weights.

\section{Datasets}
\subsection{Training Dataset}
Looking into examples from the MS-MARCO Passage Ranking \cite{msmarco} triples training set, we make the following observation: many of the passages labeled as best match for a query, contain a single span with high similarity to the query. Following this observation we view MS-MARCO as a large scale dataset with queries mapped to related spans that are surrounded by natural context. While this dataset is very large, it is also noisy, as not all positive passages are guaranteed to have a consecutive words span that is similar to the query. It is also important to note that there is no annotation in the dataset labeling what span in the matched passage is most similar to the query. We show that our method overcomes this by max-pooling from all the phrase candidates in the passage, yielding useful representations for spans.

\subsection{Evaluation Dataset}
We introduce a test set built on the Semantic Textual Similarity, STS-B \citep{cer-stsb-2017-semeval}, test dataset  that includes pairs of sentences annotated with a score between 0 and 5 for degree of similarity. We modify the dataset by generating context for the target paraphrases with the help of GPT-3.5-Turbo \citep{chagpt}. We use the following prompt:\\
\texttt{
Please provide a context to the given phrase, insert it into a long sentence, place it as part of the larger sentence and use the phrase verbatim. The sentence should also start with the letter <letter> the phrase is: <phrase>"} \\
Where \texttt{<phrase>} is replaced by the appropriate phrase from the dataset and \texttt{<letter>} is uniformly sampled from A-Z to increase diversity. 
As the text is auto generated we validate the dataset by removing examples where the generated text does not include the original phrase verbatim (23\%). We do not discard examples for having more than one sentence in the generated text. Next, we let 2 human annotators validate the remaining examples removing instances where the text generation affects the gold similarity labels (3\%), and are left with 1024 examples. We find that the results align with our experiments on private data, and share this dataset to demonstrate our findings in a reproducible fashion and accommodate future research.

\begin{table}[t]
\centering
\begin{tabular}{|| l | l | l ||}
\hline
\textbf{Model} & \textbf{Pearson} & \textbf{Spearman} \\
\hline
\multicolumn{3}{||l||}{\emph{
Baseline
} }\\
\hline
BM-25 (Okapi) & 0.342 & 0.395 \\
\hline
\multicolumn{3}{||l||}{\emph{
Full Context
} }\\
\hline
GloVe & 0.226 & 0.248 \\
BERT & 0.286 & 0.26 \\
SBERT-nli \citep{sentencebert}& 0.474 & 0.460 \\
\textbf{SBERT-msmarco} \citep{huggingfaceSentencetransformersmsmarco} & \textbf{0.557} & \textbf{0.549}\\
PhraseBERT \citep{phrasebert}& 0.518 & 0.482 \\
SLICE (ours) $\ast$ & 0.547 $\ast$ & 0.535 \\
\hline
\multicolumn{3}{||l||}{\emph{
N-grams - Forward Pass per N-gram
} }\\
\hline
BERT & 0.445 & 0.444 \\
SBERT-nli & 0.698 & 0.721 \\
SBERT-msmarco & 0.709 & 0.704 \\
PhraseBERT  & 0.731 & 0.743 \\
\textbf{SLICE (ours)}  & \textbf{0.762} & \textbf{0.757} \\
\hline
\multicolumn{3}{||l||}{\emph{
N-grams - Single Forward Pass
} }\\
\hline
GloVe & 0.452 & 0.436 \\
BERT  & 0.406 & 0.387 \\
SBERT-nli & 0.573 & 0.557 \\
SBERT-msmarco & 0.587 & 0.593 \\
PhraseBERT  & 0.643 & 0.618 \\
DensePhrases \citep{lee-etal-2021-learning-dense}& 0.562 & 0.535 \\
\textbf{SLICE (ours)} & \textbf{0.677} & \textbf{0.669} \\
\hline
\end{tabular}
\label{tab:abc}
\caption{Correlations on STS-B-Context.} Results, except when marked with $\ast$. have statistically significant difference (p<0.05) from best result in setup (marked in bold), using Williamsw t-test\citep{human-corelation-testing}, on Pearson correlation.

Only SBERT-msmarco and SLICE were trained on the same dataset (MS-MARCO)\citep{msmarco}.
\label{tab:stsvontext-corelation}
\end{table} 
\section{Experiments}
The experiments shown in Table \ref{tab:stsvontext-corelation} are separated to three different setups. The first setup, \emph{``Full Context"}, examines a single vector representing the entire context, in this case the long auto-generated sentence. The best performing model in this section is \emph{SBERT-msmarco} that was trained with an entire passage represented by a single vector. This approach yields sub-par results across all models tested.

The second setup, \emph{``N-grams - Forward per N-gram"}, produces a vector for each span in the context from single word to MAX\_SIZE of 20 words. Each such n-gram is processed with a forward pass of the model, where the input consists of the n-gram only, without its surrounding context. The space required to index all such n-grams is $\theta(NK$) where $N$ denotes all words and $K$ denotes the maximum span size. Perhaps more importantly, the required compute is significant as it executes $NK$ forward passes.

The third setup \emph{``N-grams - single forward pass''} requires that all word embeddings be acquired from \textbf{a single forward pass of the context} (i.e., sentence). We produce the representation of each n-gram in the context by mean-pooling the respective vectors of all tokens in the n-gram, with the exception of DensePhrases, in which n-grams are represented by a concatenation of the starting and ending token of the n-gram. While our trained model performs well on the other setups this is its intended use.

A heuristic is required to select the MAX\_SIZE of spans to be calculated. We select MAX\_SIZE=20 as it accounts for the maximum length (number of tokens) of 90\% of origin phrases, accommodating the scenario of a minority of phrases that exceed an assumed maximal length. Experimenting on the setup of \emph{``N-grams - single forward pass''}, with values for MAX\_SIZE, by multiples of 5 up to 35, we observe that SLICE advantage remains statistically significant between 15-35 and smaller when value is 10 or lower.

All models in these experiments (excluding GloVe and BM25) build on the  BERT-base \cite{bert} architecture, model size and  pretrained weights (except DensePhrases which is initialized from the weights of \citep{spanbert}). The models training procedures differ though, both in optimization objective and their training corpus. Our suggested method SLICE was trained on MS-MARCO similarly to SBERT-msmarco \cite{sbert-msmarco} and differs essentially in its loss function. We train for 200k steps, with a batch size of 32 which is equivalent to the amount of steps reported by \citep{sbert-msmarco}. For training, MAX\_SIZE is set to 10 based on statistics for query length in MS-MARCO. 

PhraseBERT, DensePhrases and SBERT-nli were each trained with different data. Future research can further investigate the contribution of training data versus optimization objective for these models. With respect to SBERT-msmarco, the comparison suggests a clear advantage for our proposed training objective. 

For the baseline of BM25 we use \citep{rank_bm25_github} with wikitext-103 \cite{wikitext103} for TF/IDF statistics, and for GloVe \citep{glove} we use the 300d embeddings trained on 840B tokens. 

\section{Conclusion}
We look into guided paraphrase mining and present a dataset with semantic distance scores and distracting contexts. We demonstrate the limitation of representing long contexts with a single dense vector and the effectiveness of comparing to all sub-sentence spans in the searchable text. We present the motivation for producing aggregatable token embeddings in a single encoding of the full context, and propose SLICE, a model and training objective to achieve this. We show that our proposed model outperforms alternatives of similar compute utilization. 

\bibliographystyle{ACM-Reference-Format}
\bibliography{slice-sigconf}


\begin{thebibliography}{24}


\ifx \showCODEN    \undefined \def \showCODEN     #1{\unskip}     \fi
\ifx \showDOI      \undefined \def \showDOI       #1{#1}\fi
\ifx \showISBNx    \undefined \def \showISBNx     #1{\unskip}     \fi
\ifx \showISBNxiii \undefined \def \showISBNxiii  #1{\unskip}     \fi
\ifx \showISSN     \undefined \def \showISSN      #1{\unskip}     \fi
\ifx \showLCCN     \undefined \def \showLCCN      #1{\unskip}     \fi
\ifx \shownote     \undefined \def \shownote      #1{#1}          \fi
\ifx \showarticletitle \undefined \def \showarticletitle #1{#1}   \fi
\ifx \showURL      \undefined \def \showURL       {\relax}        \fi
\providecommand\bibfield[2]{#2}
\providecommand\bibinfo[2]{#2}
\providecommand\natexlab[1]{#1}
\providecommand\showeprint[2][]{arXiv:#2}

\bibitem[Asaadi et~al\mbox{.}(2019)]%
        {bird-naacl2019}
\bibfield{author}{\bibinfo{person}{Shima Asaadi}, \bibinfo{person}{Saif~M. Mohammad}, {and} \bibinfo{person}{Svetlana Kiritchenko}.} \bibinfo{year}{2019}\natexlab{}.
\newblock \showarticletitle{Big BiRD: A Large, Fine-Grained, Bigram Relatedness Dataset for Examining Semantic Composition}. In \bibinfo{booktitle}{\emph{Proceedings of the Annual Conference of the North American Chapter of the Association for Computational Linguistics: Human Language Technologies (NAACL)}}. \bibinfo{address}{Minneapolis, USA}.
\newblock


\bibitem[Brown(2020)]%
        {rank_bm25_github}
\bibfield{author}{\bibinfo{person}{Dorian Brown}.} \bibinfo{year}{2020}\natexlab{}.
\newblock \bibinfo{booktitle}{\emph{{Rank-BM25: A Collection of BM25 Algorithms in Python}}}.
\newblock
\urldef\tempurl%
\url{https://doi.org/10.5281/zenodo.4520057}
\showDOI{\tempurl}
\newblock
\shownote{[Accessed 07-01-2024]}.


\bibitem[Cer et~al\mbox{.}(2017)]%
        {cer-stsb-2017-semeval}
\bibfield{author}{\bibinfo{person}{Daniel Cer}, \bibinfo{person}{Mona Diab}, \bibinfo{person}{Eneko Agirre}, \bibinfo{person}{I{\~n}igo Lopez-Gazpio}, {and} \bibinfo{person}{Lucia Specia}.} \bibinfo{year}{2017}\natexlab{}.
\newblock \showarticletitle{{S}em{E}val-2017 Task 1: Semantic Textual Similarity Multilingual and Crosslingual Focused Evaluation}. In \bibinfo{booktitle}{\emph{Proceedings of the 11th International Workshop on Semantic Evaluation ({S}em{E}val-2017)}}. \bibinfo{publisher}{Association for Computational Linguistics}, \bibinfo{address}{Vancouver, Canada}, \bibinfo{pages}{1--14}.
\newblock
\urldef\tempurl%
\url{https://doi.org/10.18653/v1/S17-2001}
\showDOI{\tempurl}


\bibitem[Chen et~al\mbox{.}(2017)]%
        {chen-2017-wikipedia-open-domain}
\bibfield{author}{\bibinfo{person}{Danqi Chen}, \bibinfo{person}{Adam Fisch}, \bibinfo{person}{Jason Weston}, {and} \bibinfo{person}{Antoine Bordes}.} \bibinfo{year}{2017}\natexlab{}.
\newblock \showarticletitle{Reading {W}ikipedia to Answer Open-Domain Questions}. In \bibinfo{booktitle}{\emph{Proceedings of the 55th Annual Meeting of the Association for Computational Linguistics (Volume 1: Long Papers)}}. \bibinfo{publisher}{Association for Computational Linguistics}, \bibinfo{address}{Vancouver, Canada}, \bibinfo{pages}{1870--1879}.
\newblock
\urldef\tempurl%
\url{https://doi.org/10.18653/v1/P17-1171}
\showDOI{\tempurl}


\bibitem[Devlin et~al\mbox{.}(2019)]%
        {bert}
\bibfield{author}{\bibinfo{person}{Jacob Devlin}, \bibinfo{person}{Ming-Wei Chang}, \bibinfo{person}{Kenton Lee}, {and} \bibinfo{person}{Kristina Toutanova}.} \bibinfo{year}{2019}\natexlab{}.
\newblock \bibinfo{title}{BERT: Pre-training of Deep Bidirectional Transformers for Language Understanding}.
\newblock
\newblock
\showeprint[arxiv]{1810.04805}~[cs.CL]


\bibitem[ElasticSearch(2023)]%
        {elastic-elser}
\bibfield{author}{\bibinfo{person}{ElasticSearch}.} \bibinfo{year}{2023}\natexlab{}.
\newblock \bibinfo{title}{Introducing Elastic Learned Sparse Encoder: Elastic’s AI model for semantic search}.
\newblock \bibinfo{howpublished}{\url{https://www.elastic.co/search-labs/may-2023-launch-sparse-encoder-ai-model}}.
\newblock
\newblock
\shownote{[Online; accessed 21-September-2023]}.


\bibitem[Formal et~al\mbox{.}(2021)]%
        {splade}
\bibfield{author}{\bibinfo{person}{Thibault Formal}, \bibinfo{person}{Carlos Lassance}, \bibinfo{person}{Benjamin Piwowarski}, {and} \bibinfo{person}{Stéphane Clinchant}.} \bibinfo{year}{2021}\natexlab{}.
\newblock \bibinfo{title}{SPLADE v2: Sparse Lexical and Expansion Model for Information Retrieval}.
\newblock
\newblock
\showeprint[arxiv]{2109.10086}~[cs.IR]


\bibitem[Graham and Baldwin(2014)]%
        {human-corelation-testing}
\bibfield{author}{\bibinfo{person}{Yvette Graham} {and} \bibinfo{person}{Timothy Baldwin}.} \bibinfo{year}{2014}\natexlab{}.
\newblock \showarticletitle{Testing for Significance of Increased Correlation with Human Judgment}. In \bibinfo{booktitle}{\emph{Proceedings of the 2014 Conference on Empirical Methods in Natural Language Processing ({EMNLP})}}, \bibfield{editor}{\bibinfo{person}{Alessandro Moschitti}, \bibinfo{person}{Bo~Pang}, {and} \bibinfo{person}{Walter Daelemans}} (Eds.). \bibinfo{publisher}{Association for Computational Linguistics}, \bibinfo{address}{Doha, Qatar}, \bibinfo{pages}{172--176}.
\newblock
\urldef\tempurl%
\url{https://doi.org/10.3115/v1/D14-1020}
\showDOI{\tempurl}


\bibitem[Joshi et~al\mbox{.}(2020)]%
        {spanbert}
\bibfield{author}{\bibinfo{person}{Mandar Joshi}, \bibinfo{person}{Danqi Chen}, \bibinfo{person}{Yinhan Liu}, \bibinfo{person}{Daniel~S. Weld}, \bibinfo{person}{Luke Zettlemoyer}, {and} \bibinfo{person}{Omer Levy}.} \bibinfo{year}{2020}\natexlab{}.
\newblock \showarticletitle{{S}pan{BERT}: Improving Pre-training by Representing and Predicting Spans}.
\newblock \bibinfo{journal}{\emph{Transactions of the Association for Computational Linguistics}}  \bibinfo{volume}{8} (\bibinfo{year}{2020}), \bibinfo{pages}{64--77}.
\newblock
\urldef\tempurl%
\url{https://doi.org/10.1162/tacl_a_00300}
\showDOI{\tempurl}


\bibitem[Khattab and Zaharia(2020)]%
        {colbert}
\bibfield{author}{\bibinfo{person}{Omar Khattab} {and} \bibinfo{person}{Matei Zaharia}.} \bibinfo{year}{2020}\natexlab{}.
\newblock \showarticletitle{ColBERT: Efficient and Effective Passage Search via Contextualized Late Interaction over BERT}. In \bibinfo{booktitle}{\emph{Proceedings of the 43rd International ACM SIGIR Conference on Research and Development in Information Retrieval}} (Virtual Event, China) \emph{(\bibinfo{series}{SIGIR '20})}. \bibinfo{publisher}{Association for Computing Machinery}, \bibinfo{address}{New York, NY, USA}, \bibinfo{pages}{39–48}.
\newblock
\showISBNx{9781450380164}
\urldef\tempurl%
\url{https://doi.org/10.1145/3397271.3401075}
\showDOI{\tempurl}


\bibitem[Lee et~al\mbox{.}(2021)]%
        {lee-etal-2021-learning-dense}
\bibfield{author}{\bibinfo{person}{Jinhyuk Lee}, \bibinfo{person}{Mujeen Sung}, \bibinfo{person}{Jaewoo Kang}, {and} \bibinfo{person}{Danqi Chen}.} \bibinfo{year}{2021}\natexlab{}.
\newblock \showarticletitle{Learning Dense Representations of Phrases at Scale}. In \bibinfo{booktitle}{\emph{Proceedings of the 59th Annual Meeting of the Association for Computational Linguistics and the 11th International Joint Conference on Natural Language Processing (Volume 1: Long Papers)}}. \bibinfo{publisher}{Association for Computational Linguistics}, \bibinfo{address}{Online}, \bibinfo{pages}{6634--6647}.
\newblock
\urldef\tempurl%
\url{https://doi.org/10.18653/v1/2021.acl-long.518}
\showDOI{\tempurl}


\bibitem[Merity et~al\mbox{.}(2017)]%
        {wikitext103}
\bibfield{author}{\bibinfo{person}{Stephen Merity}, \bibinfo{person}{Caiming Xiong}, \bibinfo{person}{James Bradbury}, {and} \bibinfo{person}{Richard Socher}.} \bibinfo{year}{2017}\natexlab{}.
\newblock \showarticletitle{Pointer Sentinel Mixture Models}. In \bibinfo{booktitle}{\emph{International Conference on Learning Representations}}.
\newblock
\urldef\tempurl%
\url{https://openreview.net/forum?id=Byj72udxe}
\showURL{%
\tempurl}


\bibitem[Mohammad and Hirst(2012)]%
        {mohammad-semantic-distance}
\bibfield{author}{\bibinfo{person}{Saif~M. Mohammad} {and} \bibinfo{person}{Graeme Hirst}.} \bibinfo{year}{2012}\natexlab{}.
\newblock \bibinfo{title}{Distributional Measures of Semantic Distance: A Survey}.
\newblock
\newblock
\showeprint[arxiv]{1203.1858}~[cs.CL]


\bibitem[Nguyen et~al\mbox{.}(2016)]%
        {msmarco}
\bibfield{author}{\bibinfo{person}{Tri Nguyen}, \bibinfo{person}{Mir Rosenberg}, \bibinfo{person}{Xia Song}, \bibinfo{person}{Jianfeng Gao}, \bibinfo{person}{Saurabh Tiwary}, \bibinfo{person}{Rangan Majumder}, {and} \bibinfo{person}{Li Deng}.} \bibinfo{year}{2016}\natexlab{}.
\newblock \showarticletitle{Ms marco: A human-generated machine reading comprehension dataset}.
\newblock  (\bibinfo{year}{2016}).
\newblock


\bibitem[Nogueira and Cho(2019)]%
        {bert-cross-encoding}
\bibfield{author}{\bibinfo{person}{Rodrigo~Frassetto Nogueira} {and} \bibinfo{person}{Kyunghyun Cho}.} \bibinfo{year}{2019}\natexlab{}.
\newblock \showarticletitle{Passage Re-ranking with {BERT}}.
\newblock \bibinfo{journal}{\emph{CoRR}}  \bibinfo{volume}{abs/1901.04085} (\bibinfo{year}{2019}).
\newblock
\showeprint[arXiv]{1901.04085}
\urldef\tempurl%
\url{http://arxiv.org/abs/1901.04085}
\showURL{%
\tempurl}


\bibitem[OpenAI(2023)]%
        {chagpt}
\bibfield{author}{\bibinfo{person}{OpenAI}.} \bibinfo{year}{2023}\natexlab{}.
\newblock \bibinfo{title}{Azure OpenAI Chat Completion}.
\newblock
\newblock
\urldef\tempurl%
\url{https://learn.microsoft.com/en-us/azure/ai-services/openai/concepts/models#gpt-35}
\showURL{%
Retrieved August 28, 2023 from \tempurl}


\bibitem[Pennington et~al\mbox{.}(2014)]%
        {glove}
\bibfield{author}{\bibinfo{person}{Jeffrey Pennington}, \bibinfo{person}{Richard Socher}, {and} \bibinfo{person}{Christopher Manning}.} \bibinfo{year}{2014}\natexlab{}.
\newblock \showarticletitle{{G}lo{V}e: Global Vectors for Word Representation}. In \bibinfo{booktitle}{\emph{Proceedings of the 2014 Conference on Empirical Methods in Natural Language Processing ({EMNLP})}}, \bibfield{editor}{\bibinfo{person}{Alessandro Moschitti}, \bibinfo{person}{Bo~Pang}, {and} \bibinfo{person}{Walter Daelemans}} (Eds.). \bibinfo{publisher}{Association for Computational Linguistics}, \bibinfo{address}{Doha, Qatar}, \bibinfo{pages}{1532--1543}.
\newblock
\urldef\tempurl%
\url{https://doi.org/10.3115/v1/D14-1162}
\showDOI{\tempurl}


\bibitem[Pham et~al\mbox{.}(2023)]%
        {pic-phrase-in-context-dataset}
\bibfield{author}{\bibinfo{person}{Thang Pham}, \bibinfo{person}{Seunghyun Yoon}, \bibinfo{person}{Trung Bui}, {and} \bibinfo{person}{Anh Nguyen}.} \bibinfo{year}{2023}\natexlab{}.
\newblock \showarticletitle{{P}i{C}: A Phrase-in-Context Dataset for Phrase Understanding and Semantic Search}. In \bibinfo{booktitle}{\emph{Proceedings of the 17th Conference of the European Chapter of the Association for Computational Linguistics}}. \bibinfo{publisher}{Association for Computational Linguistics}, \bibinfo{address}{Dubrovnik, Croatia}, \bibinfo{pages}{1--26}.
\newblock
\urldef\tempurl%
\url{https://aclanthology.org/2023.eacl-main.1}
\showURL{%
\tempurl}


\bibitem[Reimers and Gurevych(2019)]%
        {sentencebert}
\bibfield{author}{\bibinfo{person}{Nils Reimers} {and} \bibinfo{person}{Iryna Gurevych}.} \bibinfo{year}{2019}\natexlab{}.
\newblock \showarticletitle{Sentence-{BERT}: Sentence Embeddings using {S}iamese {BERT}-Networks}. In \bibinfo{booktitle}{\emph{Proceedings of the 2019 Conference on Empirical Methods in Natural Language Processing and the 9th International Joint Conference on Natural Language Processing (EMNLP-IJCNLP)}}. \bibinfo{publisher}{Association for Computational Linguistics}, \bibinfo{address}{Hong Kong, China}, \bibinfo{pages}{3982--3992}.
\newblock
\urldef\tempurl%
\url{https://doi.org/10.18653/v1/D19-1410}
\showDOI{\tempurl}


\bibitem["Reimers and Gurevych(2021)]%
        {huggingfaceSentencetransformersmsmarco}
\bibfield{author}{\bibinfo{person}{Nils "Reimers} {and} \bibinfo{person}{Iryna" Gurevych}.} \bibinfo{year}{2021}\natexlab{}.
\newblock \bibinfo{title}{sentence-transformers/msmarco-bert-base-dot-v5 · {H}ugging {F}ace --- huggingface.co}.
\newblock \bibinfo{howpublished}{\url{https://huggingface.co/sentence-transformers/msmarco-bert-base-dot-v5}}.
\newblock
\newblock
\shownote{[Accessed 07-01-2024]}.


\bibitem["Reimers and Gurevych(2022)]%
        {sbert-msmarco}
\bibfield{author}{\bibinfo{person}{Nils "Reimers} {and} \bibinfo{person}{Iryna" Gurevych}.} \bibinfo{year}{2022}\natexlab{}.
\newblock \bibinfo{title}{{sentence-transformers/msmarco-bert-base-dot-v5.}}
\newblock \bibinfo{howpublished}{\url{https://huggingface.co/sentence-transformers/msmarco-bert-base-dot-v5}}.
\newblock
\newblock
\shownote{[Online; accessed 11-September-2023]}.


\bibitem[Robertson and Walker(1994)]%
        {bm25}
\bibfield{author}{\bibinfo{person}{S.~E. Robertson} {and} \bibinfo{person}{S. Walker}.} \bibinfo{year}{1994}\natexlab{}.
\newblock \showarticletitle{Some Simple Effective Approximations to the 2-Poisson Model for Probabilistic Weighted Retrieval}. In \bibinfo{booktitle}{\emph{SIGIR '94}}, \bibfield{editor}{\bibinfo{person}{Bruce~W. Croft} {and} \bibinfo{person}{C.~J. van Rijsbergen}} (Eds.). \bibinfo{publisher}{Springer London}, \bibinfo{address}{London}, \bibinfo{pages}{232--241}.
\newblock
\showISBNx{978-1-4471-2099-5}


\bibitem[Wang et~al\mbox{.}(2021)]%
        {phrasebert}
\bibfield{author}{\bibinfo{person}{Shufan Wang}, \bibinfo{person}{Laure Thompson}, {and} \bibinfo{person}{Mohit Iyyer}.} \bibinfo{year}{2021}\natexlab{}.
\newblock \showarticletitle{Phrase-{BERT}: Improved Phrase Embeddings from {BERT} with an Application to Corpus Exploration}. In \bibinfo{booktitle}{\emph{Proceedings of the 2021 Conference on Empirical Methods in Natural Language Processing}}. \bibinfo{publisher}{Association for Computational Linguistics}, \bibinfo{address}{Online and Punta Cana, Dominican Republic}, \bibinfo{pages}{10837--10851}.
\newblock
\urldef\tempurl%
\url{https://doi.org/10.18653/v1/2021.emnlp-main.846}
\showDOI{\tempurl}


\bibitem[Yu and Ettinger(2020)]%
        {yu-ettinger-2020-assessing}
\bibfield{author}{\bibinfo{person}{Lang Yu} {and} \bibinfo{person}{Allyson Ettinger}.} \bibinfo{year}{2020}\natexlab{}.
\newblock \showarticletitle{Assessing Phrasal Representation and Composition in Transformers}. In \bibinfo{booktitle}{\emph{Proceedings of the 2020 Conference on Empirical Methods in Natural Language Processing (EMNLP)}}. \bibinfo{publisher}{Association for Computational Linguistics}, \bibinfo{address}{Online}, \bibinfo{pages}{4896--4907}.
\newblock
\urldef\tempurl%
\url{https://doi.org/10.18653/v1/2020.emnlp-main.397}
\showDOI{\tempurl}


\end{thebibliography}
\end{document}